\documentclass[conference]{IEEEtran}
\IEEEoverridecommandlockouts
\usepackage{cite}
\usepackage{float} 
\usepackage{amsmath,amssymb,amsfonts}

\newcommand\blfootnote[1]{%
  \begingroup
  \renewcommand\thefootnote{}\footnote{#1}%
  \addtocounter{footnote}{-1}%
  \endgroup
}

\usepackage{algorithmic}
\usepackage{graphicx}
\usepackage{textcomp}
\usepackage{flushend}
\usepackage[top=19.1mm, bottom=19.1mm,lmargin = 19.1mm, rmargin = 19.1mm]{geometry}
\usepackage{xcolor}
\def\BibTeX{{\rm B\kern-.05em{\sc i\kern-.025em b}\kern-.08em
    T\kern-.1667em\lower.7ex\hbox{E}\kern-.125emX}}
\begin{document}
\title{\vspace{0.27cm} \huge Temporal Fusion Based Mutli-scale Semantic Segmentation for Detecting Concealed Baggage Threats\\
}

\author{\IEEEauthorblockN{Muhammed Shafay, Taimur Hassan, Ernesto Damiani,  Naoufel Werghi}
\IEEEauthorblockA{Center for Cyber-Physical System (C2PS), Department of Electrical Engineering and Computer Sciences \\
Khalifa University of Science and Technology\\
Abu Dhabi, United Arab Emirates \\
\{100057573, taimur.hassan, ernesto.damiani, naoufel.werghi\}@ku.ac.ae}
}
\maketitle

\blfootnote{\noindent © 2021 IEEE.  Personal use of this material is permitted.  Permission from IEEE must be obtained for all other uses, in any current or future media, including reprinting/republishing this material for advertising or promotional purposes, creating new collective works, for resale or redistribution to servers or lists, or reuse of any copyrighted component of this work in other works. }

\begin{abstract}
Detection of illegal and threatening items in baggage is one of the utmost security concern nowadays. Even for experienced security personnel, manual detection is a time-consuming and stressful task.Many academics have created automated frameworks for detecting suspicious and contraband data from X-ray scans of luggage. However, to our knowledge, no framework exists that utilizes temporal baggage X-ray imagery to effectively screen highly concealed and occluded objects which are barely visible even to the naked eye. To address this, we present a novel temporal fusion driven multi-scale residual fashioned encoder-decoder that takes series of consecutive scans as input and fuses them to generate distinct feature representations of the suspicious and non-suspicious baggage content, leading towards a more accurate extraction of the contraband data. 
The proposed methodology has been thoroughly tested using the publicly accessible GDXray dataset, which is the only dataset containing temporally linked grayscale X-ray scans showcasing extremely concealed contraband data. The proposed framework outperforms its competitors on the GDXray dataset on various metrics.   

\end{abstract}


\section{Introduction}
\noindent Baggage threat detection is of the prime security concern over the past decade due to increased terrorism activities. X-ray imaging is one of the most commonly utilized mediums for identifying unlawful and contraband data from luggage in real-time, owing to its low operating cost and excellent visualization capabilities. However, due to increased work schedules during rush hours, it is very difficult for security staff to detect such suspicious content manually. Many academics have established autonomous baggage screening architectures for luggage to circumvent this. The initial wave of these methods was based on conventional machine learning. More recently, researchers have developed attention \cite{opixray} and contour-driven \cite{Hassan2} deep learning systems to recognize contraband data from X-ray imagery. Furthermore, many researchers handled the imbalanced nature of the threatening items in the real world by designing class imbalance resistant frameworks \cite{miao2019sixray}. Also, the problem of baggage threat detection is also addressed via semantic segmentation \cite{socpar} and instance segmentation \cite{hassan2020trainable} approaches. However, to the best of our knowledge, all of these methods are limited towards detecting cluttered suspicious objects from the single scans, and there is no framework that leverages the temporal information from X-ray imagery to detect the concealed contraband items.

\section{Related Work}
\noindent 
Baggage threat detection is a rigorously investigated topic where the academics have developed established machine learning and deep learning-based frameworks to recognize baggage threats via X-ray imagery automatically. To best present the literature, we have arranged it as per their employed machine learning and deep learning schemes.

\noindent\textbf{Traditional Approaches:} The initial baggage screening  methods employed detection \cite{Hassan2} and classification \cite{turcsany_mouton_breckon_2013} strategies using SURF and SIFT features coupled SVM model. Moreover, Franzel et al. \cite{franzel} used HOG as a feature extractor with SVM to recognize threatening items. Dai et al. \cite{dai2} used KNN to detect prohibited items via 3D reconstructed images. Heitz et al. \cite{heitz} utilized SURF characteristics in conjunction with region growing techniques to detect prohibited data in luggage X-ray images.

\noindent\textbf{Deep Learning Methods:} In this category,  researchers have utilized classification \cite{sametTL}, detection \cite{hassan2020cascaded} and segmentation \cite{socpar} approaches to identify threat objects from the security X-ray images. While many of the deep learning frameworks utilized supervised learning to detect cluttered baggage threats \cite{akcay}, researchers have also proposed semi-supervised \cite{akcay2018ganomaly}, unsupervised schemes \cite{akcay2019skipganomaly}, \cite{unsupervised_taimur} and tensor-pooling based frameworks \cite{TP} to recognize baggage threats as anomalies. Xu et al. \cite{xu_zhang_yang_2018} proposed to employ attention modules while Griffin et al. \cite{griffin} used maximum-likelihood approximations for threat object recognition. To handle the extreme occlusion, the recent approaches employ instance segmentation \cite{hassan2020trainable}, and attention mechanisms \cite{opixray} to recognize cluttered baggage threats. Still, these frameworks either have limited detection performance or require extensive parameter configurations \cite{Hassan2} to effectively recognize occluded suspicious baggage content.

\begin{figure*}[t]
\centering
    \includegraphics[width=1\linewidth]{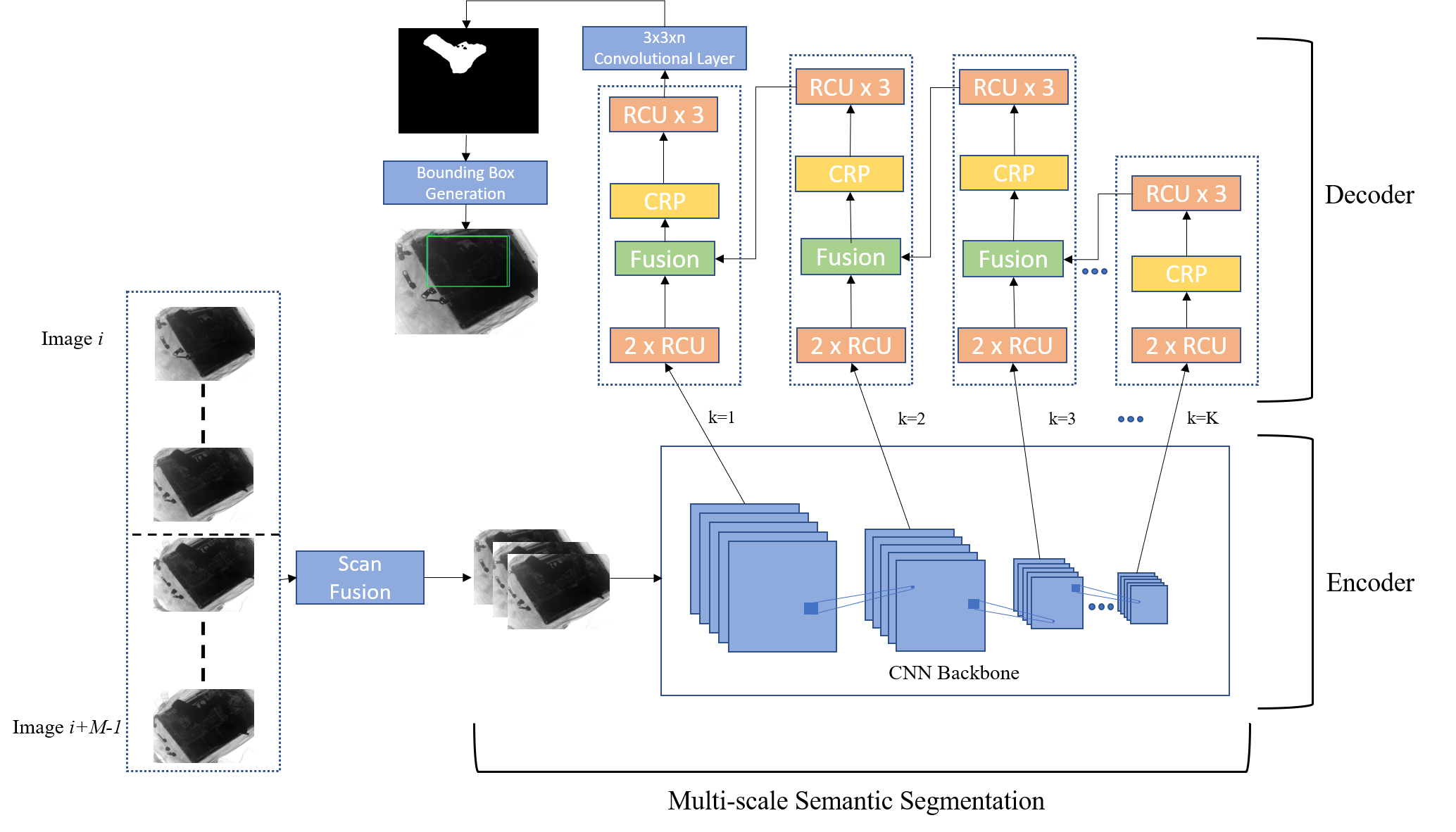}
    \caption{Block diagram of the suggested architecture. First, the consecutive $M$ input scans are fused together, and are passed to the symmetric encoder-decoder network, which generates the distinct feature representations across multiple scales using the pre-trained encoder backbone. Afterward, these features are combined together at the decoder end via custom residual convolutional units, chained residual pooling, and multi-scale fusion blocks to extract the underlying contraband data within the fused scans. $n$ in the final convolutional layer is equal to the total number of classes. Also, the mask of these extracted suspicious items is then utilized in generating their bounding boxes.}
    \label{fig:bd}
\end{figure*}

\noindent \textbf{Contributions:} 
In this study, we present a novel multi-scale fusion mechanism that utilizes temporally connected luggage X-ray images to identify highly cluttered contraband data. To our knowledge, this approach is the first attempt within the baggage threat detection domain to leverage the temporal information for recognizing extremely cluttered suspicious baggage data.The motivation for utilizing temporal information or baggage screening stems from the fact that cluttered baggage threats cannot be well-detected from the single scan, even manually by the expert staff \cite{Hassan2}. 
Hence, analyzing the single-scan limits automatic frameworks' capacity towards accurately detecting the suspicious objects. Especially in the real world scenarios, the X-ray imagery at the airport, malls, and cargoes are acquired continuously (i.e., in a consecutively manner), where single baggage is observed within multiple frames (using multiple views) \cite{bastan}. Therefore, utilizing this temporal information within the threat detection framework can potentially enhances its performance (as evident from Section \ref{sec:results}). To recapitulate, the main contributions of the paper are:         

\begin{itemize}
 \item The first baggage threat detection methodology, to our knowledge, which proposes temporal fusion in a multi-scale semantic segmentation network to effectively detect concealed baggage threats.

 \item  Extensive evaluation on publicly available GDXray dataset \cite{gdxray}, which is the only public dataset to date that provides consecutive frames of the same luggage.
\end{itemize}

\section{Proposed Methodology}
\noindent Figure \ref{fig:bd} shows the block diagram of the proposed architecture. First, we feed the proposed framework with the consecutive $M$ scans, ranging from index $i$ to $i+M-1$. These scans are temporally fused together and are passed to the encoder backbone, which generates the distinct latent representations for the suspicious baggage data. Afterward, the latent representation as well as the shallower features of the backbone network are passed to the decoder block, which reconstructs the suspicious items (across multiple scales) via residual convolutional units (RCUs), chained residual pooling (CRP), and fuses them together through addition-driven fusion mechanisms. After extracting the suspicious objects, they are localized through the bounding boxes, which are generated by analyzing each suspicious item's respective minimum and maximum values across each image dimension.   

\subsection{Scan Fusion}
\noindent When the $M$ consecutive scans of $N_1 \times N_2$ dimensions are loaded into the proposed framework, they are fused together as $N_1 \times N_2 \times 3$. In the proposed study, when $M > 3$, we always fuse the $M$ scans in pair of three as the input to the pre-trained CNN models is fixed to three-channeled. For example, when $M=9$, we will get three pairs. Moreover, when $M$ is a non-multiple of 3, we only pair based on the highest multiple of 3 and discard the remaining scans. For example, when $M=101$, we will have 99 images in the form of 33 pairs, each pair comprising of three consecutive frames, while discarding only the very first and last scans. It should be noted here the maximum number of these discarded scans (within the proposed scheme) is only two. As a result, removing two scans has little impact on the proposed system's total performance.
Apart from this, after the scan fusion process, we pass the fused representations to the custom symmetric encoder-decoder network to extract the suspicious baggage content via multi-scale semantic segmentation.

\subsection{Multi-scale Semantic Segmentation}
\noindent We propose a novel symmetric encoder-decoder topology, inspired by LWRN \cite{lwrn}, to recognize suspicious baggage items from the X-ray scans. The fused scans are first passed to the encoder backbone (having $K$ discrete feature decomposition levels). The encoder generates the latent representation of the suspicious baggage content across $k=1,2,..., K$ levels. At each level $k$, the feature representations generated by the $k^{th}$ encoder block is coupled with the corresponding decoder part via addition. Moreover, each block within the decoder part is joined together via RCU, CRP, and multi-scale fusion blocks. The total number of hyper-parameters within the proposed encoder-decoder network is 27.32M (all of these parameters are trainable). The detailed description of the units (within the decoder end) is discussed in the subsequent sections.

\begin{figure}
    \centering
    \includegraphics[width=1\linewidth]{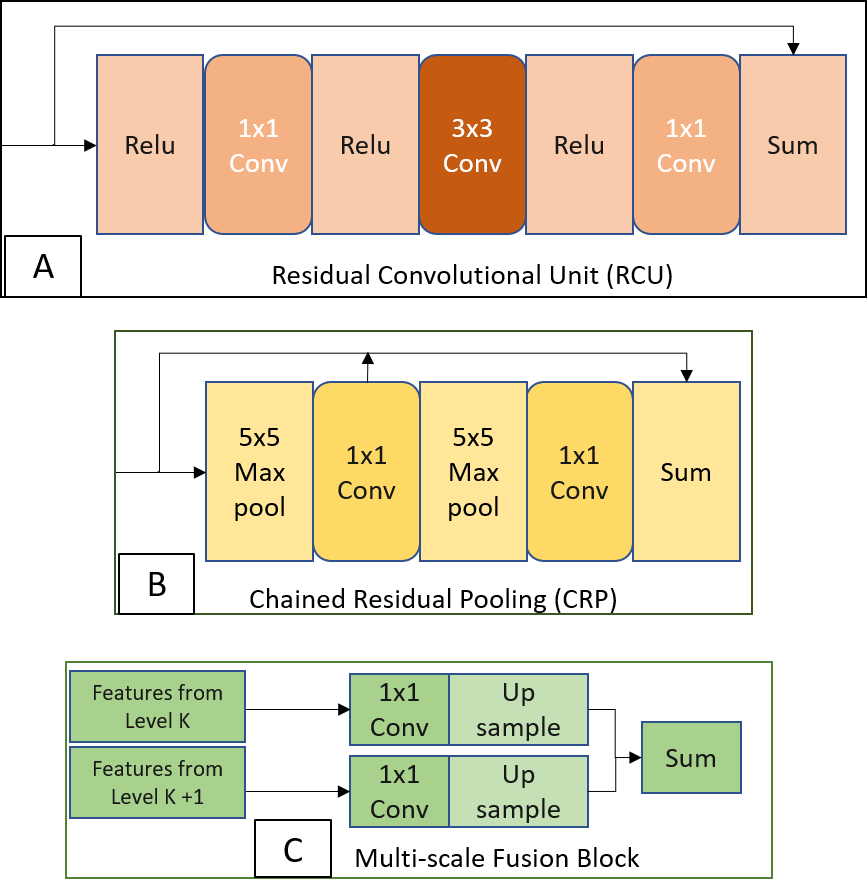}
    \caption{(A) Residual Convolutional Units, (B) Chained Residual Pooling, and (C) Multi-scale Fusion Blocks.}
    \label{fig:rcu}
\end{figure}

\subsubsection{Residual Convolutional Units (RCUs)}
\noindent RCU consists of a convolutional block called a repressed version of ResNet \cite{resnet} that lacks batch-normalization layers. RCUs are responsible for drawing out the coarser, more rough characteristics, as well as finer and softer characteristics from lower and upper blocks of the CNN backbone, as evident from Figure \ref{fig:rcu} (A), and add the residual features from the beginning to the finer features inside the RCU.

\subsubsection{Chained Residual Pooling (CRP)} 
\noindent CRP pools the distinct feature representations in a cascaded topology as shown in Figure \ref{fig:rcu} (B). Here, we can see that each CRP unit includes two times the max-pooling layer coupled with the convolutional layer, where these layers are summed together in a residual fashion. Consequently, these CRP blocks contain a vast field of view (FoV) through variable size kernels in each consecutive layer. The vast FoV allows the CRP blocks to capture a broader range of scan contextual information to predict each pixel class accurately.
\subsubsection{Multi-scale Fusion Blocks (MFB)}
\noindent The MPB units are shown in Figure \ref{fig:rcu} (C). Here, we can see that each MBP (at $j^{th}$ level) takes the feature representations from the RCU block at $k+1$ level, upscale them, and add them with the features obtained from the RCU block at the current $k^{th}$ level. Such feature fusion allows the proposed network to leverage the cluttered suspicious items' multi-scale information for their accurate extraction.  

\subsection{Bounding Box Generation}
\noindent After extracting the suspicious items, their bounding boxes are generated through a simple yet effective scheme. For each suspicious item mask, we first analyze its minimum and maximum row and column values across the candidate scan. Then, we utilize these four values to generate the bounding box as shown in Figure \ref{fig:bd}.

\section{Experimental Setup}
\noindent This section reports the comprehensive explanation of the dataset, the training,  implementation protocols as well as the assessment metrics:

\subsection{Dataset Details}
\noindent 
The proposed framework is thoroughly evaluated on the GDXray dataset \cite{gdxray} dataset. GDXray is a popular dataset that is extensively used for non-destructive testing (NDT) purposes. It is the only public dataset that consists of temporally linked grayscale X-ray scans showcasing baggage content within the consecutive frames \cite{merySurvey}. Although, the complete GDXray contains 19,407 scans for five NDT categories, we only used the baggage subset for this study as it belongs to the applicable category. The baggage category within GDXray contain 8,150 scans showcasing threat objects such as guns, razors, shuriken, and knives. It should be noted here that due to the grayscale nature of the GDXray dataset scans, it is extremely difficult to detect occluded and cluttered threat objects from the GDXray dataset, even for the expert security officers \cite{Hassan2}. 

\noindent To assess the recognition capacity of the proposed framework, we used 70\% of the images for training and 30\% for evaluation.

\subsection{Implementation Details}
\noindent To implement the framework, Python 3.7.4 with PyTorch has been employed using on a machine equipped with Intel Core i7 9300H, 16 GB RAM, and NVIDIA GTX 1650 4GB GPU with CUDA 10.1.243 and cudNN v7.5. The optimizer used during the training was stochastic gradient descent with momentum (SGDM) with a fixed learning rate of 0.0005 and a momentum of 0.9. The CNN backbone within the proposed framework is empirically determined to be ResNet-50 \cite{resnet}, whereas, in each training iteration, the proposed framework minimizes the categorical cross-entropy loss function ($l_{ce}$), as expressed below:

\begin{equation} \label{eq:L_m_new}
{l}_{ce} = -\frac{1}{b_s}\sum\limits_{u=0}^{b_s-1}\sum\limits_{v=0}^{c-1} t_{u,v}\log(p(l_{u,v})),
\end{equation}

\noindent where $b_s$ denotes he batch size, $c$ represents the total number of classes, $t_{u,v}$ denotes the ground truth labels for $u^{th}$ sample and $v^{th}$, and $p(l_{u,v})$ denotes the softmax probability of predicting the logit $l_{u,v}$ belonging to $u^{th}$ sample and $v^{th}$ class.

\begin{figure}
    \centering
    \includegraphics[width=1\linewidth]{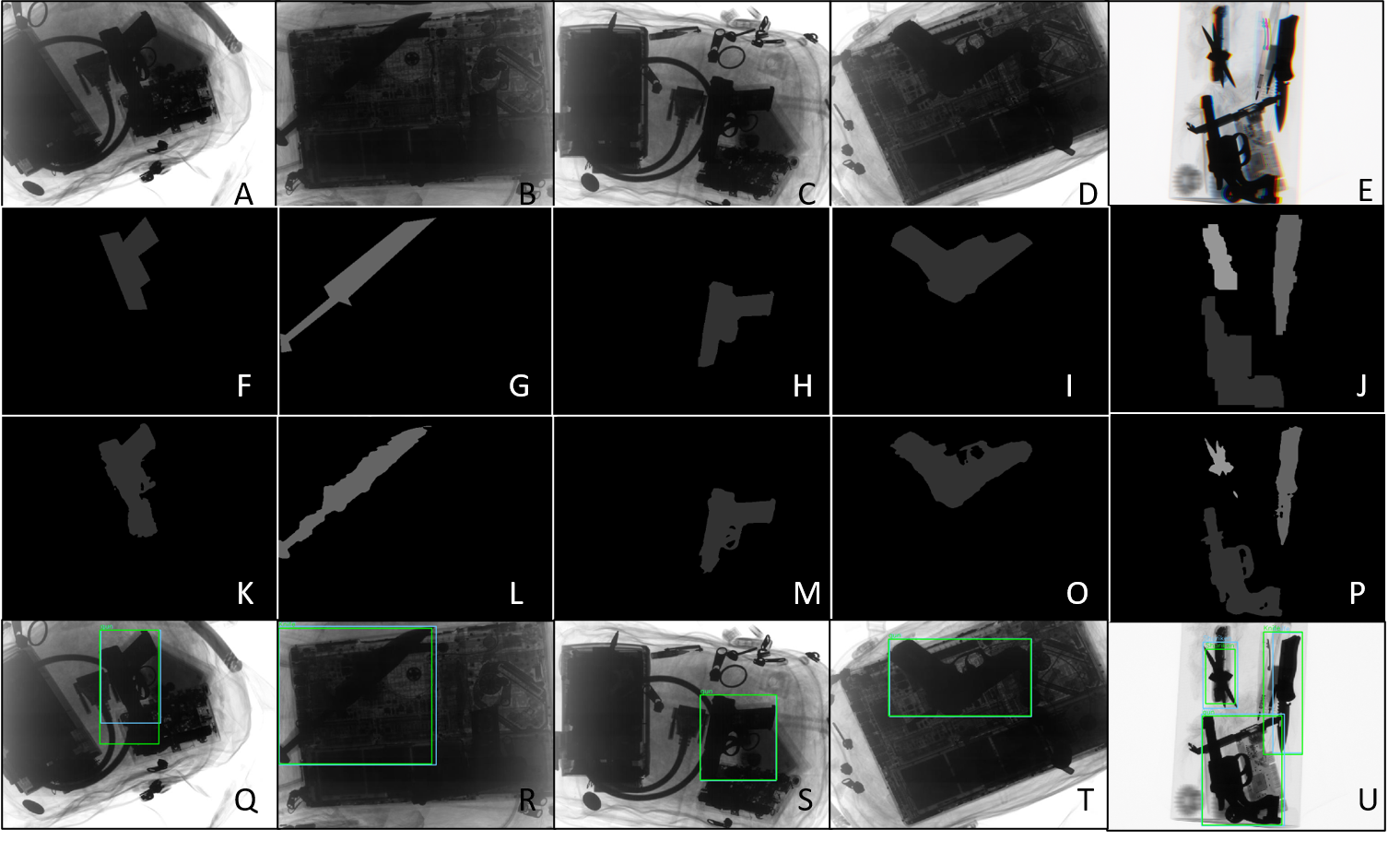}
    \caption{Quantitative evaluation of the proposed framework. (A-E) shows the original test scans, (F-J) shows the mask-level ground truth labels, (K-P) shows suspicious items extracted by the proposed framework, and (Q-U) shows the final box-level results of the proposed framework (in green color) overlaid with the ground truths (in blue color).}
    \label{fig:compQ}
\end{figure}

\begin{table}[t]
    \centering
    \caption{Performance comparison of the proposed encoder-decoder model with state-of-the-art semantic segmentation networks. Bold indicates the best scores, while underlined scores are the second best.}
    \begin{tabular}{c c c c c} 

     \hline
     Metric & Proposed & PSPNet\cite{zhao2017pspnet} & SegNet\cite{badrinarayanan2016segnet} & UNet\cite{ronneberger2015unet}\\\hline
     mIoU  & \textbf{0.9284} & 0.8836 & 0.8582 &  \underline{0.8922}\\ 
     Recall  & \textbf{0.9357} &  \underline{0.9116} & 0.8836 & 0.9086\\ 
     Precision  & 0.9587 & 0.9637 &  \underline{0.9685} & \textbf{0.9771}\\ 
     F-Score  & \textbf{0.9431} & 0.9353 & 0.9155 &  \underline{0.9382}\\ \hline
    \end{tabular}
     \label{tab:tab1}
\end{table}


\begin{table*}[t]
    \centering
    \caption{Comparison of the proposed framework with state-of-the-art baggage threat detection frameworks. Bold indicates the best performance while the second-best performance is underlined. Moreover, '-' indicates that the metric is not computed.}
     \begin{tabular}{c c c c c c c} 
     \hline
     Metric & \textbf{Proposed} &YOLOv2\cite{yolo} & Tiny YOLO\cite{yolo} & CST\cite{hassan2020cascaded} & AISM\textsubscript{SURF} \cite{aism} & AISM\textsubscript{SIFT} \cite{aism}\\ \hline
     Accuracy & \underline{0.9614}  & \textbf{0.9710} & 0.8900 & 0.9683 & - & -\\ 
     Recall  &  \textbf{0.9357} & 0.8800 & 0.8200 & \underline{0.8856} & 0.6564 & 0.8840 \\ 
     Precision  & \underline{0.9587}  & - & - & \textbf{0.9890} & 0.6300 & 0.8300 \\ 
     F-Score  & \textbf{0.9431} & 0.8996 & 0.7494 & \underline{0.9178} & - & -\\ \hline
    \end{tabular}

    \label{tab:compTable}
\end{table*}

\subsection{Evaluation Metrics}

\noindent To evaluate the performance of the proposed framework, we used the following metrics:

\subsubsection{Intersection-over-Union}
The intersection-over-union (IoU) measures the overlap among the predicted suspicious items masks and the ground truths. Mathematically, IoU is computed as $(\mathrm{IoU = \frac{T_P}{T_P + F_P + F_N}})$, where $T_P$, $F_P$, and $F_N$ represents the pixel-wise true positives, false positives, and false negatives, respectively. Afterward, we compute the mean IoU (mIoU) by averaging the IoU scores for each suspicious item category. 

\subsubsection{Recall, Precision and F-score}
To perform extensive evaluations, we also computed pixel-level recall $\mathrm{(T_{PR}=\frac{T_P}{T_P+F_N}})$, precision $\mathrm{(P_{PV}=\frac{T_P}{T_P+F_P}})$ and F-score $\mathrm{(F_1=\frac{2 x T_{PR} x P_{PV}}{T_{PR} + P_{PV}}})$.

\subsubsection{Mean Average Precision}
To measure the detection capacity of the proposed methodology, we also used the mean average precision (mAP) scores. The mAP are computed as ($mAP=\sum_{i=0}^{c-1}AP_i$), where $AP$ denotes the average precision score of each suspicious item category.

\section{Results} \label{sec:results}
\noindent In this section, the evaluation results of the proposed framework are discussed in details, as well as its comparison with the state-of-the-art approaches. To best present the results, we have categorized each set of experiment in a separate sub-section.

\subsection{Determining the Optimal Segmentation Network} 
\noindent We conducted a detailed ablation analysis in the first series of tests to assess the proposed framework's multi-scale semantic segmentation performance using state-of-the-art encoder-decoder and scene parsing networks such as PSPNet \cite{zhao2017pspnet}, SegNet \cite{badrinarayanan2016segnet}, UNet \cite{ronneberger2015unet}. The comparison is presented in Table \ref{tab:tab1}, where we can see that the proposed encoder-decoder network surpasses the second-best model by 3.51\%, 2.57\%, and 0.519\% in terms of mIoU, Recall, and F-score respectively. Although, in terms of precision, the proposed model lags 1.88\% from the second-best UNet model. But since it outperforms its competitors in all the other metrics, we chose it for conducting the rest of the experimentation.

\subsection{Comparative Analysis}
\noindent In the second set of evaluations, we compared the proposed framework's detection performance to that of state-of-the-art baggage threat detection frameworks like CST \cite{hassan2020cascaded}, DTSD \cite{dtsd}, and the YOLOv2 \cite{yolo}, and Tiny YOLO \cite{yolo} based scheme proposed in \cite{yolo}. The comparison is reported in Table \ref{tab:compTable}, where it can be seen that the proposed framework outperforms its competitors by 5.35\% in terms of recall and 2.68\% in terms of F-score. Also, it is worth noting that the proposed framework although lags from YOLOv2 \cite{yolo} by 0.988\% in terms of accuracy, it beats YOLOv2 \cite{yolo} by 4.61\% in terms of F-score, which is more promising considering the fact that accuracy is vulnerable against the imbalanced data (especially for recognizing true negatives 'background' pixels) compared to the F-score. 

\noindent Apart from this, the proposed framework also outperformed state-of-the-art occlusion-aware dual tensor-shot detector (DTSD) framework \cite{dtsd} by 2.6\% and Cascaded Structure Tensor framework \cite{hassan2020cascaded} by 1.22\% in terms of mAP scores, as evident from Table \ref{tab:map}.

\subsection{Qualitative Evaluations}
\noindent Figure \ref{fig:compQ} shows the qualitative evaluation of the proposed framework on GDXray \cite{gdxray}. In comparison to the ground truth, we can observe how effectively the proposed model retrieved the cluttered suspicious objects. Consider the findings in (R), where the cluttered knife was retrieved properly in comparison to the ground truth. Similarly, it can be observed that the proposed framework has recognized the extremely cluttered gun in (Q, S, and T) compared to the ground truth. Moreover, the quality of the masks generated by the proposed framework in (K-L) also evidences its superiority over its competitors such as CST \cite{hassan2020cascaded}, and DTSD \cite{dtsd}, which cannot generate the masks and are simply the object detection frameworks. 


\begin{table}
\centering
\caption{Comparison of the proposed framework with state-of-the-art baggage threat detectors in terms of mAP scores. Bold indicates the best performance.}
\begin{tabular}{c c c c} 
 \hline
 Metric & \textbf{Proposed} & DTSD \cite{dtsd} & CST\cite{hassan2020cascaded}\\ \hline
 Knife & \underline{0.9610} &  - & \textbf{0.9900} \\ 
 Razor & \textbf{0.9051} & - & \underline{0.8800} \\ 
 Shuriken  & \underline{0.9706} & - & \textbf{0.9900} \\ 
 Gun  & \textbf{0.9322} & - & \underline{0.9100} \\ 
 mAP & \textbf{0.9422} & 0.9162 & \underline{0.9300} \\ \hline

\end{tabular}
\label{tab:map}
\end{table}


\section{Conclusion}
\noindent In this paper, we proposed an original temporal fusion and multi-scale semantic segmentation-based baggage threat detection framework to recognize extremely concealed and cluttered contraband data from the grayscale baggage X-ray imagery. The proposed framework has been extensively compared with the state-of-the-art baggage threat detection schemes on the publicly available GDXray dataset, where it outperforms them in various metrics. In the future, we envisage testing the applicability of the proposed framework for detecting the 3D printed and organic contraband items, which, contrary to the metallic items, are barely visible within the X-ray imagery.  

\section{Acknowledgment}
\noindent This work is supported by a research fund from Khalifa University, Ref: CIRA-2019-047 and the Abu Dhabi Department of Education and Knowledge (ADEK), Ref: AARE19-156.
\bibliography{biblography}
\bibliographystyle{IEEEtran}

\end{document}